\documentclass[]{spie}  %>>> use for US letter paper
\usepackage[]{graphicx}
\usepackage[]{amsmath}
\usepackage[]{amssymb}
\usepackage[]{caption}
\usepackage[]{algorithm2e}
\usepackage[]{subcaption} 
\usepackage[]{multirow}
\usepackage[]{array}
\usepackage[]{units}
\usepackage[]{sidecap}
\usepackage{hyperref}

\DeclareMathOperator*{\argmin}{arg\,min}
\DeclareMathOperator*{\dvg}{\operatorname{div}}

\title{A Continuous Max-Flow Approach to Cyclic Field Reconstruction}

\author{John S.H. Baxter\supit{a,b}, Jonathan McLeod\supit{a,b}, and Terry M. Peters\supit{a,b}
\skiplinehalf
\supit{a}Robarts Research Institute, London, Ontario, Canada; \\
\supit{b}Western University, London, Ontario, Canada
}

\authorinfo{Send correspondence to J.S.H.B.: E-mail: jbaxter@robarts.ca}
\begin{document} 
  \maketitle 

%%%%%%%%%%%%%%%%%%%%%%%%%%%%%%%%%%%%%%%%%%%%%%%%%%%%%%%%%%%%% 
\begin{abstract}
Reconstruction of an image from noisy data using Markov Random Field theory has been explored by both the graph-cuts and continuous max-flow community in the form of the Potts and Ishikawa models. However, neither model takes into account the particular cyclic topology of specific intensity types such as the hue in natural colour images, or the phase in complex valued MRI. This paper presents \textit{cyclic continuous max-flow} image reconstruction which models the intensity being reconstructed as having a fundamentally cyclic topology. This model complements the Ishikawa model in that it is designed with image reconstruction in mind, having the topology of the intensity space inherent in the model while being readily extendable to an arbitrary intensity resolution.
\end{abstract}

%>>>> Include a list of keywords after the abstract 

%%%%%%%%%%%%%%%%%%%%%%%%%%%%%%%%%%%%%%%%%%%%%%%%%%%%%%%%%%%%%
\section{INTRODUCTION}
\label{sec:intro}
Markov random fields, specifically in the form of the Ishikawa model\cite{ishikawa2003exact}, have previously been used to address the issue of reconstructing an image from noisy data using probabilistic priors and maximum a posteriori probability optimization. The benefit of such frameworks is their submodularity and therefore global optimizability via graph cuts.\cite{kolmogorov2004energy} The primary improvement of Ishikawa models over previous graph cut approaches in reconstructing an image\cite{boykov2001fast} or depth field\cite{kolmogorov2002multi} is that the topology of the intensity space, an ordered linear topology, is fundamentally built into the functional.

Continuous max-flow approaches have traditionally been seen as continuous analogs to earlier discrete models. Along these lines, a continuous Ishikawa model \cite{bae2014fast} was proposed which maintains the same linear topology over the intensity space but using a continuous, rather than discrete domain. This is especially crucial in image reconstruction as the mitigation of discrete metrification artifacts achieved by the continuous formulation \cite{yuan2010study} has an immediate visual effect on the reconstruction.

Certain types of fields, however, do not have ordered linear intensities. For example, hue in HSL and HSV colour spaces or phase in a complex-valued image represents a cyclic, rather than linear, topology. These require a specialized reconstruction optimizer which, as the Ishikawa model was to linear intensity models, sensitive to the intensity's cyclic topology. Although this could in theory be accomplished through a directed acyclic graph continous max-flow (DAGMF) model as described by Baxter et al.,\cite{baxter2014dagmf} the quadratic growth of said models is undesirable, especially for medical applications such as phase processing in susceptibility weighted magnetic resonance imaging\cite{haacke2004susceptibility} in which high intensity resolution is necessary.

\section{Contributions}
The main contribution of this work is the development of a continuous max-flow model describing cyclic field image reconstruction, that is, reconstruction of an image in which the image intensity is defined over a cyclic topology rather than a linear topology as explored by Bae et al.\cite{bae2014fast}. This reconstruction algorithm is better suited for the reconstruction of cyclic fields such as colour/hue in natural images or phase in complex-valued magnetic resonance images.

\section{Previous Work}
Previous work by Yuan et al. \cite{yuan2010study} has addressed the continuous binary min-cut problem:
\begin{gather*}
E(u) = \int\limits_{\Omega}(D_s(x)u(x) + D_t(x)(1-u(x))+ S(x)|\nabla u(x)|)dx \\
\mbox{ s.t. } u(x) \in \{ 0, 1 \}
\end{gather*}
as well as the convex relaxed continuous Potts Model\cite{yuan2010continuous}:
\begin{gather*}
E(u) = \sum\limits_{\forall L} \int\limits_{\Omega}(D_L(x)u_L(x)+ S(x)|\nabla u_L(x)|)dx \\
\mbox{ s.t. } u_L(x) \geq 0 \mbox{ and } \sum\limits_{\forall L}u_L(x) = 1 \mbox{ .}
\end{gather*}
Both of these techniques both used a continuous max-flow model with augmented Lagrangian multipliers. In the case of the convex-relaxed continuous Potts model, the source flow had infinite capacity and the costs in the functional corresponded with constraints on the sink flows.

Bae et al. \cite{bae2014fast} extended the work on the  continuous binary min-cut problem to the continuous Ishikawa model:
\begin{gather*}
E(u) = \sum\limits_{L=0}^N \int\limits_{\Omega}(D_L(x)u_L(x)+ S(x)|\nabla u_L(x)|)dx \\
\mbox{ s.t. } u_L(x) \in \{ 0,1 \} \mbox{ and } u_{L+1}(x) \leq u_L(x)
\end{gather*}
using similar variational methods but a tiered continuous graph analogous to that used by Ishikawa \cite{ishikawa2003exact} in the discrete case, that is, with finite capacities on intermediate flows between labels.

Both continuous Potts and Ishikawa models are important to consider as they were the first max-flow models to be used in image reconstruction. However, they are not topologically suitable for cyclic fields because the Potts model disallows any notion of label topology, and the Ishikawa model is defined only over linear orderings. Additionally, cyclic orderings are non-hierarchical, making them inaccessible to generalized hierarchical max-flow solvers\cite{baxter2014ghmf} and directed acyclic graph max-flow models\cite{baxter2014dagmf} of these orderings require quadratic time to handle linear increases in the intensity resolution.

\section{Cyclic Topology in Continuous Max-Flow Segmentation}
\subsection{Continuous Max-Flow Model}
The desired optimization model for cyclic field reconstruction would be:
\begin{equation}
\underset{\theta(x)} \min \; E = \int_\Omega \Big( D_{\theta(x)}(x) + S_{\theta(x)}(x) | \nabla \theta(x) |  \Big) dx
\label{dual}
\end{equation}
in which $| \nabla \theta(x) |$ is defined over a cyclic manifold to prevent modulation errors.

One can re-express the reconstructed intensity $\theta(x)$ with respect to a series of fuzzy indicator functions:
\begin{equation*}
u_{\theta}(x) \geq 0 \text{ and } u_{\theta}(x) \approx \delta(x) \text{ iff } \theta(x) \approx \theta \text{ and }  \int_{-\pi}^\pi u_{\theta}(x) d\theta = 1
\end{equation*}
Using these indicator functions, the data term can be re-written as:
\begin{equation*}
\int_\Omega D_{\theta(x)}(x) dx = \int_{-\pi}^\pi \int_\Omega D_\theta(x) u_\theta(x) dxd\theta 
\end{equation*}
and the smoothness term can be re-written as:
\begin{equation*}
 \int_\Omega S_{\theta(x)}(x)| \nabla_{(x)} \theta(x) | dx = \int_{-\pi}^\pi \int_\Omega S_\theta(x) |\nabla_{(\theta,x)} u_\theta(x)| dxd\theta
\end{equation*}
yielding the alternative minimization formulation:
\begin{equation}
\begin{aligned}
\underset{u} \min \; E = & \int_{-\pi}^\pi \int_\Omega \left( D_\theta(x)u_\theta(x) + S_\theta(x) |\nabla u_\theta(x)|  \right) dx d\theta \\
0 \leq & u_{\theta}(x) \\
1 = & \int_{-\pi}^\pi u_{\theta}(x)d\theta \text{ and } u_{\theta}(x) \geq 0
\end{aligned}
\end{equation}
The benefit of this new formulation via indicator functions is that it resembles the previously explored continuous Potts model\cite{yuan2010continuous} with the exception that the gradient magnitude operator is defined over the Cartesian product of the spatial domain $x \in \Omega$ with a cyclic intensity domain, $\theta \in [-\pi, \pi)$. Thus, the spatial domain augmented with the $\theta$ direction forms a high dimensional cylinder as shown in Figure \ref{fig:cylinder}.

\subsection{Primal Formulation}
\begin{SCfigure}
\centering
\includegraphics[width=0.4\textwidth]{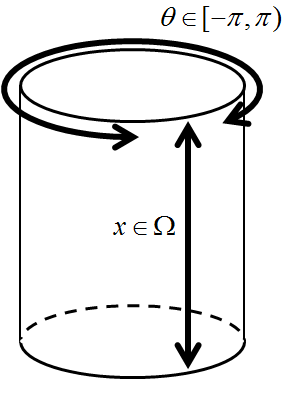}
\caption{Topology used in cyclic max-flow in which the spatial domain, $\Omega$ is one dimensional. The resulting topology has one more dimensional than $\Omega$ which encodes the intensity $\theta$ for the spatial flows, $q_\theta(x)$, and indicator functions, $u_\theta(x)$. The gradient magnitude and divergence operators over $q_\theta(x)$ and  $u_\theta(x)$ include both the spatial and the $\theta$ dimensions.}
\label{fig:cylinder}
\end{SCfigure}

As in the previous work, the primal formulation will be expressed as a max-flow optimization, specifically:
\begin{equation}
\label{primal-first}
\begin{aligned}
\underset{p,q,\lambda} \max & \int_\Omega p_S(x) dx
\end{aligned}
\end{equation}
subject to the capacity constraints:
\begin{equation*}
\begin{aligned}
p_\theta(x) & \leq D_\theta(x), \\
|q_\theta(x)| & \leq S_\theta(x) 
\end{aligned}
\end{equation*}
and the flow conservation constraint:
\begin{equation*}
0 = G_\theta(x) =\operatorname{div} q_\theta(x) + p_\theta(x) - p_S(x) \text{ .}
\end{equation*}
$q_\theta(x)$, the spatial flow now has an additional dimension, the $\theta$ direction. The $\theta$ direction is equipped with a cyclic topology and therefore is not susceptible to modulation artifacts. Similar to $u_\theta(x)$, $q_\theta(x)$ makes use of the spatial domain augmented with the cyclic domain of $\theta$ as shown in Figure \ref{fig:cylinder}.

\subsection{Primal-Dual Formulation}
The primal model can be converted to a primal-dual model through the use of Lagrangian multipliers on the flow conservation constraint $G_\theta(x) = \operatorname{div} q_\theta(x) + p_\theta(x) - p_S(x) = 0$. This yields the equation:
\begin{equation}
\label{lagrangian}
\begin{aligned}
\underset{u} \min \underset{p,q} \max & \left( \int_\Omega p_S(x) dx +  \int_{-\pi}^\pi \int_\Omega u_{\theta}(x) G_\theta(x) dx d\theta \right) \\
p_\theta(x) & \leq D_\theta(x) \\
|q_\theta(x)| & \leq S_\theta(x)
\end{aligned}
\end{equation}

To ensure that this function meets the criteria of the minimax theorem, we must ensure that it is convex with respect to $u$, considering $p,q$ to be fixed, and concave with respect to $p,q$ with $u$ fixed. \cite{ekeland1976convex} For the first, it is sufficient to note that if $p,q$ are fixed, then $G$ is fixed as well, meaning that \eqref{lagrangian} is linear and therefore convex with respect to $u$. It should also be noted that $G$ is a linear function of $p,q$, implying the linearity (and concavity) of \eqref{lagrangian} with respect to $p,q$. Thus, a saddle point must exist and the formulations are equivalent. \cite{ekeland1976convex}

\subsection{Dual Formulation}
To show the equivalence of the primal-dual formulation (Eq. \eqref{lagrangian}) and the desired dual formulation (Eq. \eqref{dual}), we must show that the original formula can be reconstructed from the definition of the saddle-point. First, consider the primal-dual formulation with the sink flows, $p_\theta(x)$ isolated:
\begin{equation}
\underset{u} \min \underset{p_\theta(x) \leq D_\theta(x)} \max \int_\Omega u_\theta(x)p_\theta(x)dx = \underset{u_\theta(x) \geq 0} \min  \int_\Omega u_\theta(x)D_\theta(x)dx
\end{equation}
which implies that $u_\theta(x) \geq 0$ as if $u_\theta(x) < 0$, the minimization is unbounded as $p_\theta(x) \to -\infty$. This reconstructs the data term portion of the original formulation using the fuzzy labeling function.

The source flow, $p_S$ in Eq. \ref{lagrangian} can be isolated as:
\begin{equation}
\underset{u} \min \underset{p_S(x)} \max \left( \int_\Omega p_S(x)dx - \int_{-\pi}^\pi  \int_\Omega u_\theta(x)p_S(x) dx \right) = 0
\end{equation}
at the saddle point defined by $1 =  \int_{-\pi}^\pi u_\theta(x) d\theta$. This reconstructs the labeling constraint, that is, that only one intensity value is assigned to each voxel.

Considering coupling the spatial domain $\Omega$ with the cyclic domain of $\theta$ via a simple Cartesian product with members taking the form  $z=\left[ \begin{smallmatrix}x \\ \theta \end{smallmatrix} \right]$. In this coupled space, the maximization of the spatial flow functions can be expressed \cite{giusti1984minimal} as:
\begin{equation} \label{eq:regularSpatialFlow}
\begin{aligned}
\int_{\Omega \times \theta \in [\pi, \pi] } u_\theta(x) \dvg q_\theta(x)dz
& = \int_{\Omega \times \theta \in [\pi, \pi] } \left( \dvg (u_\theta(x)q_\theta(x)) - q_\theta(x) \cdot \nabla u_\theta(x) \right) dz \\
& = \int_{\Omega \times \theta \in [\pi, \pi] } \dvg (u_\theta(x)q_\theta(x)) dx  - \int_{\Omega \times \theta \in [\pi, \pi] }  q_\theta(x) \cdot \nabla u_\theta(x) dz \\
& = \int_{\theta \in [\pi, \pi]}\oint_{\delta\Omega} u_\theta(x)q_\theta(x) \cdot d\mathbf{s}d\theta  - \int_{\Omega \times \theta \in [\pi, \pi] }  q_\theta(x) \cdot \nabla u_\theta(x) dz \\
& = - \int_{\Omega \times \theta \in [\pi, \pi] }  q_\theta(x) \cdot \nabla u_\theta(x) dz \\
\underset{|q_\theta| \leq S_\theta(x)} \max \int_\Omega u_\theta(x) \dvg q_\theta(x)dz
& = \underset{|q_\theta| \leq S_\theta(x)} \max  - \int_\Omega  q_\theta(x) \cdot \nabla u_\theta(x) dz \\
& =  - \int_{\Omega \times \theta \in [\pi, \pi] }  \left( -\frac{S_\theta(x)}{|\nabla u_\theta(x)|} \nabla u_\theta(x) \right) \cdot \nabla u_\theta(x) dz \\
& = \int_{\Omega \times \theta \in [\pi, \pi] } S_\theta(x) |\nabla u_\theta(x)| dz \\
\underset{u_\theta \geq 0} \min \underset{|q_\theta| \leq S_\theta(x)} \max \int_{\Omega \times \theta \in [\pi, \pi] } u_\theta(x) \dvg q_\theta(x)dz 
& = \underset{u_\theta \geq 0} \min \int_{\Omega \times \theta \in [\pi, \pi] } S_\theta(x) |\nabla u_\theta(x)| dz  \\
& = \underset{u_\theta \geq 0} \min  \int_\Omega S_{\theta(x)}(x) |\nabla \theta(x)| dx \text{ .}
\end{aligned}
\end{equation}
This yields the smoothness component in Eq. \eqref{dual} thus showing the equivalence of the two models.

\subsection{Solution to Primal-Dual Formulation}
The optimization problem being addressed is the augmented form of Eq. \eqref{lagrangian}:
\begin{equation}
\label{augmented}
\begin{aligned}
\underset{u} \min \underset{p,q} \max & \left( \int_\Omega p_S(x) dx +  \int_{-\pi}^\pi \int_\Omega u_{\theta}(x) G_\theta(x) dx d\theta - \int_{-\pi}^\pi \int_\Omega \frac{c G_\theta^2 (x)}{2} dx d\theta \right) \\
p_\theta(x) & \leq D_\theta(x) \\
|q_\theta(x)| & \leq S_\theta(x)
\end{aligned}
\end{equation}
in which $c$ is a non-negative constant which has no effect on Eq. \eqref{lagrangian} when the constraint $G\theta(x)=0$ holds, but serves as an additional penalty. Optimizing Eq. \eqref{augmented} can be achieved by performing the following steps iteratively, optimizing Eq. \eqref{augmented} with respect to each individual variable:
{
\begin{enumerate}
\setlength{\belowdisplayskip}{2pt} \setlength{\belowdisplayshortskip}{2pt}
\setlength{\abovedisplayskip}{2pt} \setlength{\abovedisplayshortskip}{2pt}
\item Maximize \eqref{augmented} over $q_\theta$:
\begin{equation*}
q_\theta(x) \gets \operatorname{Proj}_{|q_\theta(x)| \leq S_\theta(x) } \left( q_\theta(x) + \tau \nabla \left( \dvg q_\theta(x) + p_\theta(x) - p_S(x) - u_\theta(x)/c \right) \right) 
\end{equation*}
which is a Chambolle projection iteration with descent parameter $\tau > 0$.\cite{chambolle2004algorithm} Note that both the divergence and gradient operators are evaluated over both the linear spatial domain and the cyclic $\theta$ domain.
\item Maximize \eqref{lagrangian} over the sink flows, $p_\theta$, by:
\begin{equation*}
p_\theta(x) \gets \min\lbrace D_\theta(x), p_S(x) - \dvg q_\theta(x) + u_\theta(x)/c \rbrace
\end{equation*}
\item Maximize \eqref{lagrangian} over the source flow $p_S$ analytically by:
\begin{equation*}
p_S(x) \gets \frac{1}{2\pi} \left( 1/c + \int_{-\pi}^\pi \left( p_\theta(x) + \dvg q_\theta(x) - u_\theta(x)/c \right)d\theta \right)
\end{equation*}
\item Minimize \eqref{lagrangian} over $u_\theta$ analytically by:
\begin{equation*}
u_\theta(x) \gets u_\theta(x) - c \left( \dvg q_\theta(x) - p_S(x) + p_\theta(x) \right)
\end{equation*}
\end{enumerate}
}

The specific augmented Lagrangian based algorithm used is:
\begin{algorithm}[!h]
$\forall x, \theta, u_\theta(x) = \nicefrac{1}{2\pi}$ \;
\While{not converged} {
  $\forall (x,\theta), q_\theta(x) \gets \operatorname{Proj}_{|q_\theta(x)| \leq S_\theta(x) } \left( q_\theta + \tau \nabla \left( \dvg q_\theta(x) + p_\theta(x) - p_S(x) - u_\theta(x)/c \right) \right)$\;
    $\forall (x,\theta), p_\theta(x) \gets \min \{ D_\theta(x), p_S(x)-\dvg q_\theta(x) + u_\theta(x) / c \}$\;
  $p_S(x) \gets \frac{1}{2\pi}(1/c + \int_{-\pi}^{\pi} (p_\theta(x)+\dvg q_\theta(x) - u_\theta(x)/c ) d\theta$\;
    $\forall (x,\theta), u_\theta(x) \gets u_\theta(x) - c \left( \dvg q_\theta(x) - p_S(x) + p_\theta(x) \right)$\;
}
\label{alg:solverSeq}
\end{algorithm}

\subsection{Pseudo-Flow Formulation}
The approach taken here for the use of proximal Bregman or pseudo-flow methods is derived from Baxter et al.\cite{baxter2015proximal}. The corresponding non-smooth pseudo-flow model can be written as:
\begin{equation}
\max \limits_{ |q_\theta(x)| \leq S_\theta(x) } \int_{\Omega} \min_{\theta \in [-\pi,\pi]} \left( D_\theta(x) + \dvg q_\theta(x) \right) dx
\label{pseudo}
\end{equation}
which can be derived from Eq \eqref{lagrangian} by:
\begin{equation*}
\begin{aligned}
&\min \limits_{u_\theta(x)} \max \limits_{p_S(x), q(x), p_\theta(x)}  \int_{\Omega} \left( p_S(x) + \int_{-\pi}^\pi u_\theta(x) \left( \dvg q_\theta(x) - p_S(x) + p_\theta(x) \right) \right) d\theta dx \\
& \mbox{ s.t. }  p_\theta(x) \leq D_\theta(x), \; |q_\theta(x)| \leq S_\theta(x) \\
= & \min \limits_{u_\theta(x)} \max \limits_{p_S(x), q(x), p_\theta(x)} \int_{\Omega} \left( p_S(x) - \int_{-\pi}^\pi u_\theta(x) p_S(x)d\theta + \int_{-\pi}^\pi u_\theta(x) \left( \dvg q_\theta(x) + p_\theta(x) \right) d\theta \right) dx \\
& \mbox{ s.t. } p_\theta(x) \leq D_\theta(x), \; |q_\theta(x)| \leq S_\theta(x) \\
= & \min \limits_{u_\theta(x)} \max \limits_{q_\theta(x), p_\theta(x)} \int_{\Omega} \int_{-\pi}^\pi u_\theta(x) \left( \dvg q_\theta(x) + p_\theta(x) \right) d\theta dx \\
& \mbox{ s.t. } \int_{-\pi}^\pi u_\theta(x) d\theta = 1, \; p_\theta(x) \leq D_\theta(x), \; |q_\theta(x)| \leq S_\theta(x)\\
& \mbox{ (as the equivalence of equations \eqref{dual} and \eqref{lagrangian} guarantees the constraint $\int_{-\pi}^\pi u_\theta(x) d\theta = 1$ )} \\
=& \min \limits_{u_\theta(x)} \max \limits_{q(x)} \int_{\Omega} \int_{-\pi}^\pi u_\theta(x) \left( \dvg q_\theta(x) + D_\theta(x) \right) d\theta dx \\
& \mbox{ s.t. }\int_{-\pi}^\pi u_\theta(x) d\theta = 1, \; u_\theta(x) \geq 0, \; |q_\theta(x)| \leq S_\theta(x)\\
& \mbox{ (as the equivalence of equations \eqref{dual} and \eqref{lagrangian} guarantees the constraint $u_\theta(x) \geq 0$  and } \\
& \mbox{ $\max_{p_\theta(x) \leq D_\theta(x)}u_\theta(x)p_\theta(x) = D_\theta(x)$ if $u_\theta(x)$ is non-negative) }\\
= &\max \limits_{q_\theta(x)} \min \limits_{u_\theta(x)} \int_{\Omega} \int_{-\pi}^\pi u_\theta(x) \left( \dvg q_\theta(x) + D_\theta(x) \right) dx \\
& \mbox{ s.t. } \int_{-\pi}^\pi u_\theta(x)d\theta = 1, \; u_\theta(x) \geq 0, \; |q_\theta(x)| \leq S_\theta(x)\\
= & \max \limits_{ |q_\theta(x)| \leq S_\theta(x) } \int_{\Omega} \min_\theta \left( D_\theta(x) + \dvg q_\theta(x) \right) dx \\
& \mbox{(since $\min_{u_\theta(x)} \int_{-\pi}^\pi u_\theta(x)\left( D_\theta(x) + \dvg q_\theta(x) \right) = \min_\theta ( D_\theta(x) + \dvg q_\theta(x) )$ when}\\
& \mbox{ $\int_{-\pi}^\pi u_\theta(x)d\theta = 1$ and $ u_\theta(x) \geq 0$)}
\end{aligned}
\end{equation*}

Now that a pseudo-flow representation is developed, we can take advantage of Bregman proximal projections\cite{bregman1967relaxation} to optimize this formula. Consider the distance between labeling functions $u_\theta(x)$ and $v_\theta(x)$ as:
\begin{equation}
d_g(u,v) = \int_\Omega \int_{-\pi}^\pi \left( u_\theta(x) \ln ( u_\theta(x) / v_\theta(x) ) - u_\theta(x) + v_\theta(x) \right) d\theta dx
\end{equation}
which can be verified to be a Bregman distance (when $u_\theta(x) \in [0,1]$) using the entropy function:
\begin{equation*}
g(u) = \int_\Omega \int_{-\pi}^\pi \left( u_\theta(x) \ln u_\theta(x) - u_\theta(x) \right) d\theta dx \mbox{ .}
\end{equation*}

If we consider a feasible labeling, $v_\theta(x)$, we can find another proximal labeling, $u_\theta(x)$, which has a lower energy by addressing the optimization:
\begin{equation}
u_\theta(x) = \argmin\limits_{u_\theta(x) \geq 0, \, \int_{-\pi}^\pi u_\theta(x)d\theta = 1} \left(  \max \limits_{ |q_\theta(x)| \leq S_\theta(x) } \int_{\Omega} u_\theta(x) \left( D_\theta(x) + \dvg q_\theta(x) \right) dx + c d_g(u,v) \right)
\label{bregmanopt}
\end{equation}
where $c$ is a positive constant. Using a Lagrangian multiplier on the constraint $\int_{-\pi}^\pi u_\theta(x) d\theta = 1$, we can solve for $u(x)$ analytically as:
\begin{equation}
u_\theta(x) = \frac{v_\theta(x) \exp \left( - \frac{D_\theta(x) + \dvg q_\theta(x) }{c} \right)}{ \int_{-\pi}^\pi v_{\theta'}(x) \exp \left( - \frac{D_{\theta'}(x) + \dvg q_{\theta'}(x) }{c} \right)d\theta'}
\label{labelupdate}
\end{equation}
noting that this answer fulfills the constraint $u_\theta(x) \geq 0$ provided the same holds for $v_\theta(x)$. By letting the distance weighting parameter $c$ approach 0, $u_\theta(x) \to 0$ if $D_\theta(x) + \dvg q_\theta(x) \neq \min_{\theta'} ( D_{\theta'}(x) + \dvg q_{\theta'}(x) )$. Using that fact,
\begin{equation*}
\begin{aligned}
 & \min\limits_{u_\theta(x) \geq 0, \, \int_{-\pi}^\pi u_\theta(x) d\theta = 1} \left(  \max \limits_{ |q_\theta(x)| \leq S_\theta(x) } \int_{\Omega} u_\theta(x) \left( D_\theta(x) + \dvg q_\theta(x) \right) dx + c d_g(u,v) \right) \\
\to & \max \limits_{ |q_\theta(x)| \leq S_\theta(x) } \int_\Omega \min\limits_{\theta \in [-\pi,\pi]}\left( D_\theta(x) + \dvg q_\theta(x) \right) \mbox{ as } c \to 0 &
 \end{aligned}
\end{equation*}
illustrating that the Bregman proximal method is a smoothed version of the non-smooth equation Eq \eqref{pseudo}.

\subsection{Solution to Pseudo-Flow Formulation}
By taking the gradient of equation \eqref{bregmanopt} with respect to $\dvg q_\theta(x)$ (with $u_\theta(x)$ substituted by equation \eqref{labelupdate}) one can derive the appropriate Chambolle iteration scheme\cite{chambolle2004algorithm} for maximizing \eqref{bregmanopt} with respect to the spatial flow variables $q_\theta(x)$:
\begin{equation}
q_\theta(x) \gets \operatorname{Proj}_{|q_\theta(x)| \leq S_\theta(x)} \left( q_\theta - c\tau \nabla \left(v_\theta(x) \exp \left(- \frac{D_\theta(x) + \dvg q_\theta(x)}{c} \right) \right) \right)
\label{flowupdate}
\end{equation}
where $\tau$ is a positive gradient descent parameter. Thus, the proximal Bregman, or pseudo-flow, algorithm relies on the alternation between the analytic optimization of the label functions, $u_\theta(x)$ through Eq. \eqref{labelupdate} and the specified Chambolle iteration.

The proximal Bregman based algorithm proposed is therefore:
\begin{algorithm}[!h]
$\forall x, \theta, u_\theta(x) = \nicefrac{1}{2\pi}$ \;
\While{not converged} {
  $\forall (x,\theta), q_\theta(x) \gets \operatorname{Proj}_{|q_\theta(x)| \leq S_\theta(x)} \left( q_\theta - c\tau \nabla \left(u_\theta(x) \exp \left(- \frac{D_\theta(x) + \dvg q_\theta(x)}{c} \right) \right) \right)$ \;
  $\forall (x,\theta), u_\theta(x) \gets u_\theta(x) \exp \left( - \frac{D_\theta(x) + \dvg q_\theta(x) }{c} \right)$\;
  $\forall (x,\theta), u_\theta(x) \gets \frac{u_\theta(x)}{\int_{-\pi}^\pi u_\theta(x) d\theta}$\;
}
\label{alg:solverSeq}
\end{algorithm}

\section{Discussion and Conclusions}
In this paper, we present two algorithms for addressing continuous max-flow image reconstruction in which the intensity being reconstructed is fundamentally organized along a cyclic, as opposed to linear, topology. Such a reconstruction method is more efficient than that developed from previous continuous max-flow methods capable of incorporating said topology, specifically DAGMF\cite{baxter2014dagmf}, by displaying linear rather than quadratic growth in complexity in terms of the intensity resolution. This solver has been implemented using GPGPU acceleration due to its inherent parallelizability. This solver is available open-source for both two- and three-dimensional images at \url{www.advancedsegmentationtools.org}. 

Future work in continuous max-flow based image reconstruction includes the robust automatic detection of gradient direction which may allow for anisotropic smoothness terms to be employed. These terms could be used to estimate level sets in the image, maintaining them while allowing for faster or less-constrained gradients in the direction orthogonal to said level set.

%%%%%%%%%%%%%%%%%%%%%%%%%%%%%%%%%%%%%%%%%%%%%%%%%%%%%%%%%%%%%
\acknowledgments %>>>> equivalent to \section*{ACKNOWLEDGMENTS}
The authors would like to acknowledge Zahra Hosseini and Dr.\ Martin Rajchl for their invaluable discussion and editing.

%%%%%%%%%%%%%%%%%%%%%%%%%%%%%%%%%%%%%%%%%%%%%%%%%%%%%%%%%%%%%
%%%%% References %%%%%

\bibliographystyle{spiebib}   %>>>> makes bibtex use spiebib.bst
\bibliography{TechReportCyclicMaxFlow} 

\begin{thebibliography}{10}

\bibitem{ishikawa2003exact}
H.~Ishikawa, ``Exact optimization for markov random fields with convex
  priors,'' {\em Pattern Analysis and Machine Intelligence, IEEE Transactions
  on}~{\bf 25}(10), pp.~1333--1336, 2003.

\bibitem{kolmogorov2004energy}
V.~Kolmogorov and R.~Zabin, ``What energy functions can be minimized via graph
  cuts?,'' {\em Pattern Analysis and Machine Intelligence, IEEE Transactions
  on}~{\bf 26}(2), pp.~147--159, 2004.

\bibitem{boykov2001fast}
Y.~Boykov, O.~Veksler, and R.~Zabih, ``Fast approximate energy minimization via
  graph cuts,'' {\em Pattern Analysis and Machine Intelligence, IEEE
  Transactions on}~{\bf 23}(11), pp.~1222--1239, 2001.

\bibitem{kolmogorov2002multi}
V.~Kolmogorov and R.~Zabih, ``Multi-camera scene reconstruction via graph
  cuts,'' in {\em Computer Vision—ECCV 2002},  pp.~82--96, Springer, 2002.

\bibitem{bae2014fast}
E.~Bae, J.~Yuan, X.-C. Tai, and Y.~Boykov, ``A fast continuous max-flow
  approach to non-convex multi-labeling problems,'' in {\em Efficient
  Algorithms for Global Optimization Methods in Computer Vision},
  pp.~134--154, Springer, 2014.

\bibitem{yuan2010study}
J.~Yuan, E.~Bae, and X.-C. Tai, ``A study on continuous max-flow and min-cut
  approaches,'' in {\em Computer Vision and Pattern Recognition (CVPR), 2010
  IEEE Conference on},  pp.~2217--2224, IEEE, 2010.

\bibitem{baxter2014dagmf}
J.~S. Baxter, M.~Rajchl, J.~Yuan, and T.~M. Peters, ``A continuous max-flow
  approach to multi-labeling problems under arbitrary region regularization,''
  {\em arXiv preprint arXiv:1405.0892} , 2014.

\bibitem{haacke2004susceptibility}
E.~M. Haacke, Y.~Xu, Y.-C.~N. Cheng, and J.~R. Reichenbach, ``Susceptibility
  weighted imaging (swi),'' {\em Magnetic Resonance in Medicine}~{\bf 52}(3),
  pp.~612--618, 2004.

\bibitem{yuan2010continuous}
J.~Yuan, E.~Bae, X.-C. Tai, and Y.~Boykov, ``A continuous max-flow approach to
  potts model,'' in {\em Computer Vision--ECCV 2010},  pp.~379--392, Springer,
  2010.

\bibitem{baxter2014ghmf}
J.~S. Baxter, M.~Rajchl, J.~Yuan, and T.~M. Peters, ``A continuous max-flow
  approach to general hierarchical multi-labeling problems,'' {\em arXiv
  preprint arXiv:1404.0336} , 2014.

\bibitem{ekeland1976convex}
I.~Ekeland and R.~Temam, ``Convex analysis and 9 variational problems,'' 1976.

\bibitem{giusti1984minimal}
E.~Giusti, {\em Minimal surfaces and functions of bounded variation}, no.~80,
  Springer Science \& Business Media, 1984.

\bibitem{chambolle2004algorithm}
A.~Chambolle, ``An algorithm for total variation minimization and
  applications,'' {\em Journal of Mathematical imaging and vision}~{\bf
  20}(1-2), pp.~89--97, 2004.

\bibitem{baxter2015proximal}
J.~S. Baxter, M.~Rajchl, J.~Yuan, and T.~M. Peters, ``A proximal bregman
  projection approach to continuous max-flow problems using entropic
  distances,'' {\em arXiv preprint arXiv:1501.07844} , 2015.

\bibitem{bregman1967relaxation}
L.~M. Bregman, ``The relaxation method of finding the common point of convex
  sets and its application to the solution of problems in convex programming,''
  {\em USSR computational mathematics and mathematical physics}~{\bf 7}(3),
  pp.~200--217, 1967.

\end{thebibliography}

\end{document}